\documentclass[sn-mathphys,Numbered]{sn-jnl}

\usepackage{graphicx}%
\usepackage{multirow}%
\usepackage{amsmath,amssymb,amsfonts}%
\usepackage{amsthm}%
\usepackage{mathrsfs}%
\usepackage[title]{appendix}%
\usepackage{xcolor}%
\usepackage{textcomp}%
\usepackage{manyfoot}%
\usepackage{booktabs}%
\usepackage{algorithm}%
\usepackage{algorithmicx}%
\usepackage{algpseudocode}%
\usepackage{listings}%
\usepackage{gensymb}
\usepackage[font=small,labelsep=space]{caption}
\usepackage{hyperref}

\makeatletter
\def\UrlAlphabet{%
      \do\a\do\b\do\c\do\d\do\e\do\f\do\g\do\h\do\i\do\j%
      \do\k\do\l\do\m\do\n\do\o\do\p\do\q\do\r\do\s\do\t%
      \do\u\do\v\do\w\do\x\do\y\do\z\do\A\do\B\do\C\do\D%
      \do\E\do\F\do\G\do\H\do\I\do\J\do\K\do\L\do\M\do\N%
      \do\O\do\P\do\Q\do\R\do\S\do\T\do\U\do\V\do\W\do\X%
      \do\Y\do\Z}
\def\UrlDigits{\do\1\do\2\do\3\do\4\do\5\do\6\do\7\do\8\do\9\do\0}
\g@addto@macro{\UrlBreaks}{\UrlOrds}
\g@addto@macro{\UrlBreaks}{\UrlAlphabet}
\g@addto@macro{\UrlBreaks}{\UrlDigits}
\makeatother

\theoremstyle{thmstyleone}%

\theoremstyle{thmstyletwo}%

\theoremstyle{thmstylethree}%

\begin{document}

\title[Article Title]{A sustainable development perspective on urban-scale roof greening priorities and benefits}

\author[1,3,4]{\fnm{Jie} \sur{Shao}} 

\author*[1,2]{\fnm{Wei} \sur{Yao}}\email{wei.hn.yao@polyu.edu.hk}

\author[3,4]{\fnm{Lei} \sur{Luo}} 

\author[1]{\fnm{Linzhou} \sur{Zeng}} 

\author[1]{\fnm{Zhiyi} \sur{He}} 

\author[1]{\fnm{Puzuo} \sur{Wang}} 

\author[3,4]{\fnm{Huadong} \sur{Guo}}

\affil[1]{\orgdiv{Department of Land Surveying and Geo-Informatics}, \orgname{The Hong Kong Polytechnic University}, \orgaddress{\state{Hong Kong}}}


\affil[2]{\orgdiv{Otto Poon Charitable Foundation Smart Cities Research Institute}, \orgname{The Hong Kong Polytechnic University}, \orgaddress{\city{Hong Kong}}}

\affil[3]{\orgname{International Research Center of Big Data for Sustainable Development Goals}, \orgaddress{\city{Beijing}, \postcode{100094}, \country{China}}}

\affil[4]{\orgdiv{Aerospace Information Research Institute}, \orgname{Chinese Academy of Sciences}, \orgaddress{\city{Beijing}, \postcode{100094}, \country{China}}}


\abstract{Greenspaces are tightly linked to human well-being. Yet, rapid urbanization has exacerbated greenspace exposure inequality and declining human life quality. Roof greening has been recognized as an effective strategy to mitigate these negative impacts. Understanding priorities and benefits is crucial to promoting green roofs. Here, using geospatial big data, we conduct an urban-scale assessment of roof greening at a single building level in Hong Kong from a sustainable development perspective. We identify that 85.3\% of buildings reveal potential and urgent demand for roof greening. We further find green roofs could increase greenspace exposure by \textasciitilde61\% and produce hundreds of millions (HK\$) in economic benefits annually but play a small role in urban heat mitigation (\textasciitilde0.15\degree{C}) and annual carbon emission offsets (\textasciitilde0.8\%). Our study offers a comprehensive assessment of roof greening, which could provide reference for sustainable development in cities worldwide, from data utilization to solutions and findings.}

\maketitle

\section{Introduction}\label{sec1}

Urbanization is a global megatrend of the 21st century. An increasing number of people are relocating to urban areas from rural ones, and the United Nations predicted that by 2030, 60 percent of the global population will reside in cities\cite{MANSO2021110111}. While continuous urbanization can promote economic development, the expansion and population growth of cities place immense strain on the environment and contribute to climate change\cite{cuthbert2022global, wooster2022urban}. The conversion of natural surfaces into high-density buildings and impermeable materials disrupts the balance of greenspaces, leading to increased occurrences of floods, urban heat island (UHI) effects, and higher energy consumption\cite{MANSO2021110111,chen2022contrasting,chung2021natural,shafique2020photovoltaic}. Furthermore, human activities result in higher levels of carbon emissions and urban heat\cite{shafique2020photovoltaic}. These issues adversely affect the public’s physical and mental health, producing increased mortality and morbidity\cite{wong2021greenery,maes2021benefit}. In response to unsustainable urban living, the United Nations specified to make cities and human settlements inclusive, safe, resilient, and sustainable in the Sustainable Development Goal 11 “Sustainable Cities and Communities”\cite{desa2016transforming}. Achieving this goal is a pressing challenge faced by the global community as urbanization continues to accelerate.

Greenspaces, which are urban nature areas designed to mitigate the negative impacts of urbanization and safeguard public health\cite{tost2019neural}, have become a focus of Sustainable Development Goal 11. In highly urbanized cities, building roofs make up a significant portion of impervious surfaces\cite{ASADI20201846} and present an opportunity to create additional greenspaces. Green roofs, also known as vegetated roofs, are covered with various types of plants and growing substrates, garnering widespread attention in this regard\cite{SHAFIQUE2018757}. Green roofs are commonly classified into intensive and extensive types. Intensive green roofs are characterized by a thicker substrate layer, facilitating a wide variety of plants as shrubs and small trees. This leads to more requirements on high load-bearing structures and regular maintenance. Extensive green roofs have a shallower substrate layer and are often vegetated with grasses and wildflowers, with lighter weight, lower maintenance costs, and more common utilization than intensive green roofs\cite{MANSO2021110111, FLECK2022108673}. Both green roof types can enhance the urban ecosystem and biodiversity by increasing the available greenspaces\cite{DENG201720, wooster2022urban}, possess the ability to control peak flow and runoff through stormwater retention in plant and substrate layers\cite{MANSO2021110111, FLECK2022109274}, and mitigate the UHI effects by increasing evapotranspiration, providing shade, and increasing albedo\cite{wong2021greenery, FLECK2022108673}, thereby bringing co-benefits in terms of energy saving, carbon sequestration\cite{FLECK2022109703}, and environment\cite{IRGA2022109712}. Given the positive effects of green roofs, many countries have legislated to promote their use in cities\cite{LIBERALESSO2020104693}. To ensure judicious greening policies, consideration is needed on how to prioritize green roofs and what benefits can be offered by these green roofs.

The evaluation of roof greening priority typically involves three steps\cite{XU2021108392}. Firstly, potential green roofs are identified based on building attributes. These attributes were commonly obtained through on-site measurements or government records\cite{SILVA2017120}, which limited the evaluation to a small scale\cite{HONG2019102468}. Satellite remote sensing images enable urban-scale evaluation, though they often overlook the need for roof greening on an individual building level\cite{zhang2023carbon}. Secondly, appropriate greening indicators are selected and measured using building attributes collected through remote sensing and surveys\cite{GRUNWALD201754}. These indicators help establish criteria for determining roof greening priorities. However, green roofs are known to be associated with multiple factors. Therefore, relying solely on building attributes and specific benefits, such as a combination of building height and air pollution reduction\cite{VIECCO2021108120} and a combination of building locations and runoff reduction\cite{LIU2022130064}, is insufficient for objectively prioritizing them. Lastly, roof greening priorities are calculated by combining these greening indicators. Currently, most existing studies focus on discussing specific cases\cite{HE2020122205, SHAFIQUE2020119471}, and there is a lack of comprehensive studies that assess roof greening priorities and benefits at an urban scale from the individual building level. These shortcomings may lead to a subjective understanding of the role of green roofs in sustainable urban development, thereby hampering the formulation of reasonable greening policies.

Due to low maintenance costs and universality, in this study, extensive green roofs are comprehensively assessed in Hong Kong, a thoroughly urbanized city with a population density of 22,297 people per square kilometer\cite{demographia2023}, and herbaceous plants were chosen as the preferred plant species for roof greening. Despite a high proportion of greenspaces (about 0.7) across Hong Kong, the greenspaces are uneven, where human exposure to greenspace index value is only 0.35\cite{chen2022contrasting}. Hong Kong has a monsoon-influenced humid subtropical climate (Köppen climate classification Cwa), particularly hot and rainy during the summer, with an average annual rainfall of around 2,000 mm (\url{https://i-lens.hk/hkweather/}), along with high-density buildings and abundant hard impervious surfaces, contributing to the UHI effects, flooding, and increased carbon emissions. Consequently, the contradiction between urbanization and human well-being is increasingly fierce in Hong Kong. Thus, from a sustainable urban development perspective, we used urban geospatial big data to conduct a comprehensive and refined evaluation of roof greening priorities and benefits in Hong Kong, overcoming the limitations of previous small-scale single-cases rough evaluations. Specifically, instead of solely assessing roof greening priorities based on building attributes, we first utilized building geometry and attribute information to identify potential green roofs, then focused on greening indicators closely related to urban sustainability and quantified them to determine the urgency of roof greening. By combining all indicators, we were able to evaluate the greening priority of each building. Furthermore, a multifaceted benefit of green roofs was assessed. This study allows a reproducible method for prioritizing urban-scale green roofs at a single building level and could serve as a valuable reference for city planners and decision-makers.

\section{Results}\label{sec2}

\subsection{Roof greening priorities}\label{sec21}

\begin{figure}[t]
\centering
\includegraphics[width=0.8\linewidth]{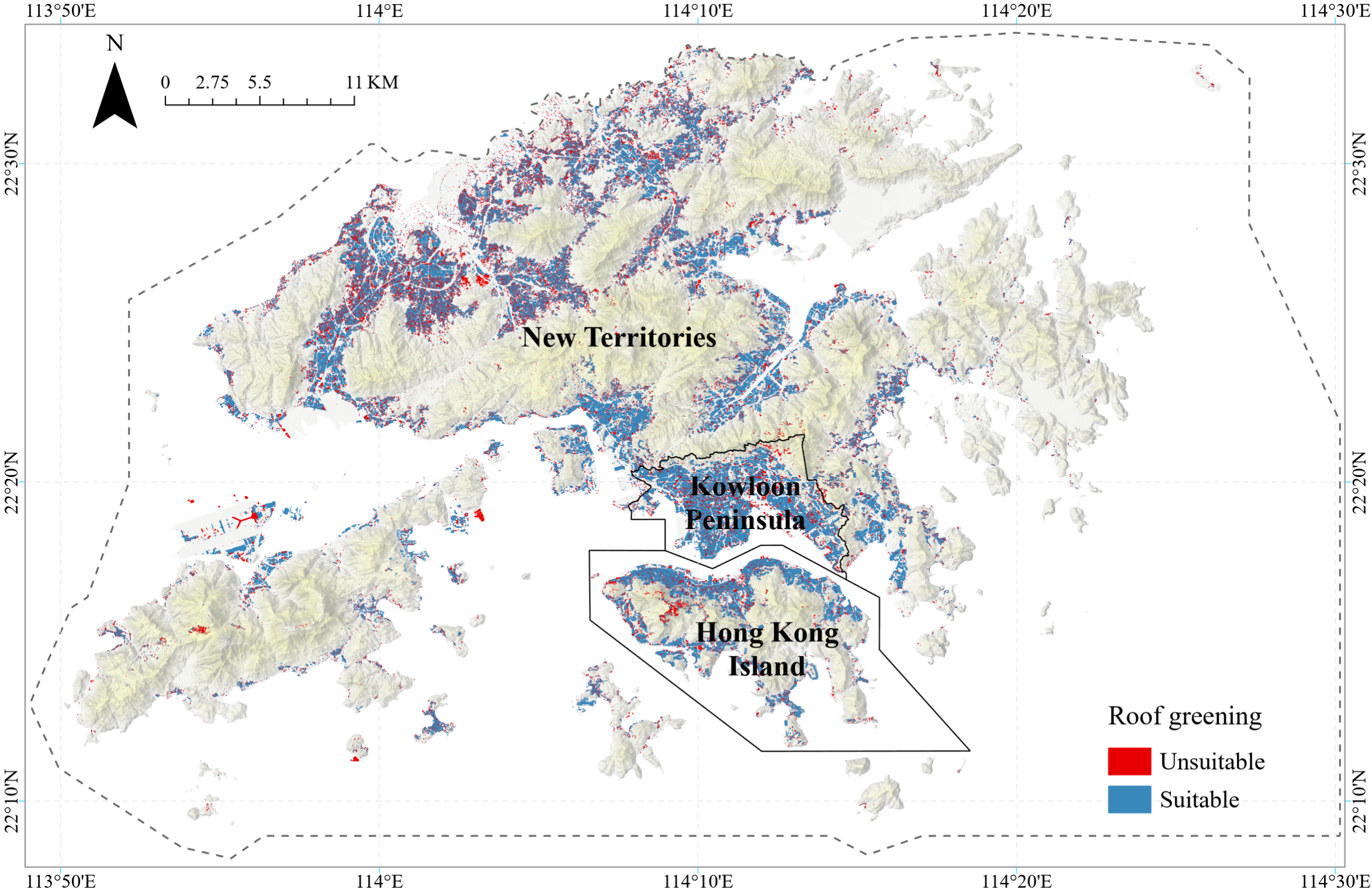}
\caption{\textbf{$\vert$ Potential green roofs in Hong Kong.} A combination of roof geometric properties (slope and area) and building age was used to identify potential green roofs, where buildings with roof slopes exceeding 15 degrees, areas smaller than 10 m$^2$, or those older than 60 years are deemed unsuitable for greening retrofit.}
\label{fig:1}
\end{figure}

\textbf{Potential green roofs.} Roof slope, roof area, and building age were used to identify potential green roofs in Hong Kong. Given the rich spatial information, airborne laser scanning point clouds obtained from an online database (Common Spatial Data Infrastructure Portal, \url{https://portal.csdi.gov.hk/csdi-webpage/}) were used to gather roof geometric properties. The results reveal that 85.3\% of buildings in Hong Kong can be retrofitted with greenery, and the area of potential green roofs amounted to 63.9 km$^2$, which accounts for around 90\% of the total roof area (70.1 km$^2$) (Fig.~\ref{fig:1}). This indicates the vast potential for roof greening in Hong Kong.

\vspace{1em}

\begin{figure}[t]
\centering
\includegraphics[width=1\linewidth]{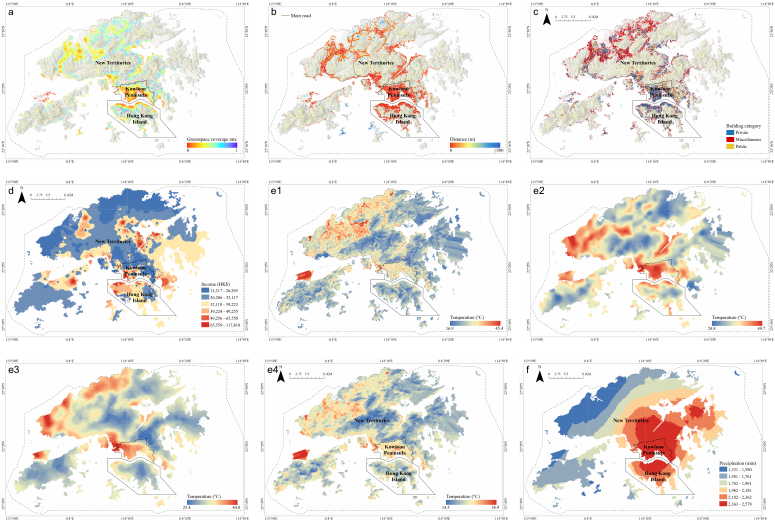}
\caption{\textbf{$\vert$ Roof greening indicators in Hong Kong.} \textbf{a}, Greenspace coverage rate within a 500-meter horizontal radius of buildings. \textbf{b}, The horizontal distance of the building to the nearest main road. \textbf{c}, Distribution of building categories. \textbf{d}, Median monthly income distribution by interpolating the income of 540 housing estates and all districts in 2021. \textbf{e1-e4}, The surface temperature by season in 2021 derived from multi-temporal Landsat-8 images (\textbf{e1}: spring; \textbf{e2}: summer; \textbf{e3}: autumn; \textbf{e4}: winter). \textbf{f}, Precipitation distribution in Hong Kong obtained by interpolating records from 139 weather stations in 2021.}
\label{fig:2}
\end{figure}

\noindent\textbf{Roof greening indicators.} Six indicators related to the environment (greenspace coverage rate and distance), social economy (building category and income), and climate (surface temperature and annual precipitation), were designed to evaluate roof greening priority. Urban areas with dense buildings, such as the Kowloon Peninsula and Hong Kong Island, typically display low greenspace coverage rates, often falling below 0.2 (Fig.~\ref{fig:2}a). This underscores the demand for increasing greenery in these areas. Conversely, the rate increases to 0.5 or even higher as the area extends towards the outskirts bordering forests, leading to a diminished demand for greenery. The distribution of distance (Fig.~\ref{fig:2}b) reveals that most buildings are located close to main roads, highlighting the demand for reducing exhaust gas and noise from vehicles, particularly in urban areas. The distribution of building categories (Fig.~\ref{fig:2}c) indicates that the majority of buildings in Hong Kong belong to the private or miscellaneous categories. Private buildings are concentrated in the Kowloon Peninsula and Hong Kong Island, resulting in a low category indicator value in these regions. Income distribution (Fig.~\ref{fig:2}d) shows that a significant portion of high-income groups reside near the coastline. Assuming that greenspaces have a positive effect on increasing income, the demand for roof greening should be higher in areas away from the coastline. Fig.~\ref{fig:2}e reveals seasonal surface temperatures, with built-up areas experiencing higher temperatures compared to vegetated suburbs and remote regions, highlighting the UHI effect. Summer and autumn experience higher temperatures, justifying a higher weight for greening. Furthermore, Hong Kong has substantial annual rainfall, particularly during the summer. Fig.~\ref{fig:2}f indicates that the urban areas in the Kowloon Peninsula and Hong Kong Island receive significantly more rainfall than those in the New Territories. Notably, the Kowloon Peninsula, characterized by dense buildings and hard impervious surfaces, received over 2,360 mm of precipitation in 2021. This high level of rainfall increases the likelihood of urban flooding, necessitating a greater emphasis on roof greening to reduce peak flow and runoff.

\vspace{1em}

\begin{figure}[b!]
\centering
\includegraphics[width=1\linewidth]{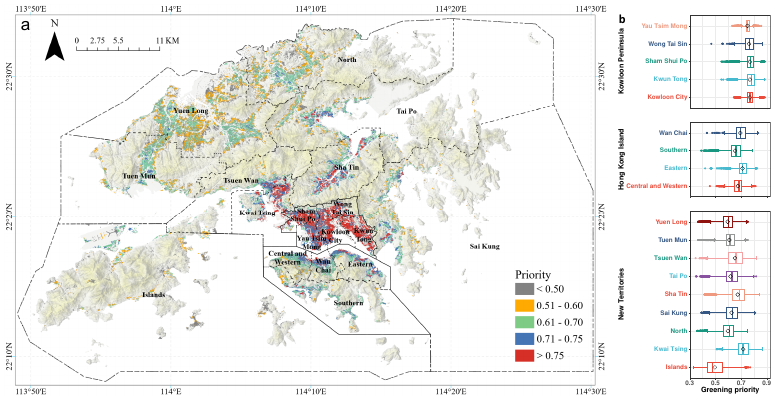}
\caption{\textbf{$\vert$ Roof greening priorities in Hong Kong.} \textbf{a}, Distribution of roof greening priority, the greening priority is almost all above 0.7 in Kowloon Peninsula, varies between 0.6 and 0.75 in Hong Kong Island, and mainly distributes below 0.7 in New Territories. \textbf{b}, Boxplot depicting the distribution of priority values across each district.}
\label{fig:3}
\end{figure}

\noindent\textbf{Roof greening priorities}. Equal weights were assigned to each nondimensionalized indicator to evaluate the greening priorities. Our results reveal that the greening priorities of most potential roofs (91\%) were greater than 0.5, the maximum priority reached about 0.9, and the average priority was above 0.6, highlighting urgent needs for roof greening in Hong Kong. Typically, roofs in urban areas tend to have higher greening priority than those in suburban areas (Fig.~\ref{fig:3}a). There are several reasons for this disparity. Urban areas have a low greenspace coverage rate, resulting in a greater need for greenery. Additionally, due to the presence of numerous impervious surfaces, sparse greenspaces, and high levels of anthropogenic activities, residents in urban areas are exposed to higher thermal stress. Consequently, the demand for urban heat reduction is higher. Conversely, suburban areas, especially those heavily vegetated, are less affected by heatwaves and therefore have lower demand for roof greening. Furthermore, urban areas typically have more main roads and higher traffic flows, necessitating additional green spaces to mitigate the adverse impact of vehicular activities like exhaust gas and noise. On the other hand, the greening priority is unequal among all districts. The priority in built-up areas, such as the Kowloon Peninsula and Hong Kong Island, is generally higher. For instance, the average priority is as high as 0.76 in the Kowloon Peninsula, particularly in Sham Shui Po and Kwun Tong Districts (Fig.~\ref{fig:3}b), and 0.67 in Hong Kong Island. In contrast, the New Territories, which consist of more suburban areas, have a priority average of 0.6. Despite the presence of more private buildings in the Kowloon Peninsula and Hong Kong Island, which typically have lower category indicator values, these areas have sparse greenspaces and complex road networks. Moreover, precipitations in the Kowloon Peninsula and Hong Kong Island are greater than those in New Territories. Interestingly, there are no significant differences in resident income and surface temperature between the Kowloon Peninsula and the New Territories. Hong Kong Island has higher income levels and lower surface temperatures, leading to lower indicator values for these factors. Consequently, buildings in built-up areas, especially in the Kowloon Peninsula, exhibit higher greening priorities after incorporating all the indicators.

\subsection{Roof greening benefits}\label{sec22}
The benefits of green roofs on four sustainable urban development goals, including ecological environment, urban heat mitigation, carbon emission reduction, and social economy, were evaluated in this study.

\vspace{1em}

\noindent\textbf{Benefits on environment.} Hong Kong has a significant amount of greenspace, comprising approximately 70\% of its land, with only 7\% designated for residential use. Most of the greenspaces, such as woodland and shrubland, are located in remote regions that are inaccessible to residents\cite{LAU2021127251}. On the other hand, residential land is concentrated in urban areas, resulting in low average coverage of greenspace around buildings, approximately 0.35 (Supplementary Fig. 1a). The coverage rate is lower in urban centers like Kowloon Peninsula and Hong Kong Island, even below 0.1 (Fig.~\ref{fig:2}a). Due to the high population density, the human exposure to greenspace index is only 35.3\%, which is much lower compared to some developed countries and regions\cite{chen2022contrasting}. This contrast between the abundant greenspace and limited human exposure has made it necessary for city planners to consider changing the status quo of uneven greenspaces. Roof greening can increase the greenspace area by 63.9 km$^2$, which is about 9\% of the existing greenspace area in Hong Kong\cite{JIM201665}. Although this may seem relatively small, it has significant implications for urban ecological environments. Firstly, green roofs can raise the average greenspace coverage rate to 0.58 (Supplementary Fig. 1b), effectively enhancing greenery provision in urban areas. Secondly, green roofs can increase human exposure to greenspace, with the potential to reach a globally high level of 56.7\%. This not only improves the overall eco-environment quality but also has the potential to promote human health and well-being. Moreover, large-scale green roofs are expected to stimulate greater insect growth through the progressively dense and intricate vegetation structure and attract a wider variety of bird species, while at the same time, birds and wind could introduce more plant species\cite{DENG201720}, thus contributing to the enrichment and conservation of urban biodiversity.

\vspace{1em}

\begin{figure}[t]
\centering
\includegraphics[width=1\linewidth]{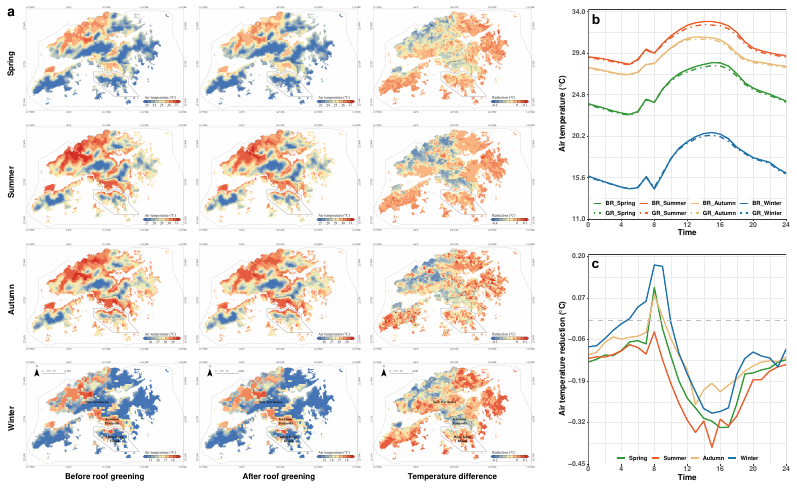}
\caption{\textbf{$\vert$ Simulation of air temperature at 2 meters above the ground for the four seasons.} \textbf{a}, The distribution of air temperature before and after roof greening, as well as their difference. \textbf{b}, The daily temperature variation under bare roof (BR) and green roof (GR) conditions. \textbf{c}, Temperature reduced by green roofs.}
\label{fig:4}
\end{figure}

\noindent\textbf{Urban heat mitigation.} To evaluate the cooling effect of green roofs in Hong Kong, the Weather Research and Forecasting/Urban Canopy Model (WRF/UCM) model was used to simulate the air temperatures at 2 m above the ground throughout the four seasons under bare roof and green roof conditions\cite{skamarock2019description}. Fig.~\ref{fig:4}a reveals that urban areas experienced higher temperatures compared to the suburban areas covered by dense vegetation. Additionally, the high-temperature areas in the Kowloon Peninsula and Hong Kong Island were smaller than those in the New Territories. These findings were consistent with the results derived from satellite images (Fig.~\ref{fig:2}e). After roof greening, the air temperature in urban areas decreased by 0-0.4\degree{C}. The high-temperature areas also diminished, with the greatest reduction observed in built-up areas, particularly in the New Territories (Fig.~\ref{fig:4}a). When examining the daily temperature variation (Fig.~\ref{fig:4}b), it was observed that winter was the coldest season, with temperatures below 20\degree{C}, thus requiring no cooling measures. Spring exhibited high air temperatures compared to winter, but generally low cooling demands. Conversely, summer and autumn recorded high air temperatures, indicating the demand for cooling solutions. The peak temperature reduction occurred between 13:00 and 16:00, when ground radiation and evaporation from porous surfaces were at their highest, resulting in greater temperature reduction. Due to surface heat loss and less evaporation, the cooling effect was least effective around 8:00 a.m., and there was even a warming effect in seasons other than summer (Fig.~\ref{fig:4}c). Overall, the cooling effect of green roofs is not significant in Hong Kong.

\vspace{1em}

\noindent\textbf{Carbon emission reduction.} Green roofs can offset carbon emissions through two patterns. Firstly, carbon dioxide can be directly absorbed by plant photosynthesis\cite{SHAFIQUE2020119471}. We estimated that implementing extensive green roofs could sequester approximately 93,000 tons $\text{CO$^2$}\cdot\text{yr}^{-1}$. Secondly, green roofs can indirectly reduce carbon emissions through energy savings\cite{ZHENG2023128018}. The estimation indicates that the indirect pattern could result in a reduction of approximately 183,000 tons $\text{CO$^2$}\cdot\text{yr}^{-1}$. Thus, the total carbon emissions reduced by green roofs is around 276,000 tons $\text{CO$^2$}\cdot\text{yr}^{-1}$. This represents about 0.8\% of annual carbon emissions in Kong Hong (about 34.7 million tons of carbon dioxide equivalent in 2021, \url{https://cnsd.gov.hk/en/climate-ready/ghg-emissions-and-trends/}). Therefore, while green roofs can contribute to Hong Kong’s carbon-neutral goal, additional efforts will be needed to fully realize this goal.

\vspace{1em}

\noindent\textbf{Benefits on social economy.} The socio-economic benefits of green roofs are often indirectly assessed through energy savings and carbon pricing\cite{TEOTONIO2021102781}. Green roofs can mitigate urban heat, thereby reducing the need for cooling and electricity consumption. According to the assumed period of cooling throughout summer and autumn, the estimated energy savings can amount to approximately 2.33 $\times{10}^8\text{kW}\cdot$h, and the cost savings can amount to around HK\$300 million, based on an electricity tariff of 1.29 HK\$$\cdot\left(\text{kW}\cdot\text{h}\right)^{-1}$ unit in 2023. Moreover, with the push towards carbon-neutral goals, carbon trading has become an effective incentive and low-cost scheme for reducing global carbon emissions. Taking the reference transaction price of China’s carbon market in June 2023 (about 65 HK\$$\cdot\text{ton}^{-1}$), the estimated monetary value of carbon emission reduction benefits from green roofs is about HK\$18 million per year. Overall, roof greening could provide an annual value of around HK\$318 million based on energy savings and carbon emission reduction benefits. These economic benefits should be considered when making decisions about roof greening projects.

\section{Discussion}\label{sec3}

Building age serves as a critical metric in assessing the feasibility of roof greening, often exhibiting a correlation with the load-bearing capacity of buildings\cite{XU2021108392, HONG2019102468}. As the building ages, the load-bearing capacity of its roof tends to progressively diminish, reaching a point where it becomes unsuitable for retrofitting with greenery\cite{LIU2022130064}. This underscores the potential for roof greening in emerging cities or regions. However, building age information (Supplementary Fig. 2) indicates that a significant portion of Hong Kong's buildings was constructed between the 1960s and 1980s, which raises concerns about roof stability and mechanical resistance. Consequently, while there is a pressing need for roof greening in many cities, city planners must carefully navigate the challenges posed by the aging condition of buildings.

Green roofs typically lower rooftop surface temperatures by several degrees, sometimes exceeding ten degrees\cite{wong2021greenery, FLECK2022108673}. However, the rooftop surface temperature does not accurately reflect thermal comfort at the pedestrian level. Instead, analyzing the air temperature at 1.5-2 m above the ground provides a more representative metric of urban thermal comfort and is suitable for assessing the energy-saving benefits of green roofs\cite{ABOELATA2021119514}. Microclimate simulation tools like ENVI-met\cite{JAMEI2023110965} are commonly employed to analyze the reduction in air temperature attributable to green roofs but prove challenging to apply at the urban scale. The WRF, a mesoscale model, offers the capability to simulate the urban-scale heat mitigation effects of green roofs\cite{skamarock2019description}. Comparison with ground measurements across four seasons reveals that the root mean square error of air temperature simulated by WRF is approximately 1\degree{C}, with a correlation of around 0.9 (Supplementary Fig. 3), showing an accuracy comparable to related studies\cite{WANG2022109082}. From Fig.~\ref{fig:4} and Supplementary Fig. 4, there is a discernible correlation between building height and density and the cooling effect. For instance, the cooling effect of green roofs in areas with high-density and high-rise buildings, such as West Kowloon and Kong Kong Island, is markedly less pronounced than in areas with low-rise buildings, such as the New Territories, to the extent that it is ineffective for human thermal comfort near the ground. This observation aligns with the findings of previous research\cite{NG2012256}. However, despite factors such as building height limiting the cooling effect of green roofs at ground level, this does not negate the fact that a cooling effect still exists. In this context, estimating energy savings based on this cooling effectiveness becomes feasible. EnergyPlus software is often used to simulate the energy-saving benefits\cite{ABOELATA2021119514}. However, these methods necessitate intricate three-dimensional models detailing floors, floor layouts, door and window orientation and placement, building materials, and green roof composition\cite{BEVILACQUA2021111523}. This complexity restricts their applicability to individual buildings or small-scale zones, rendering them impractical for big cities with tens of thousands of buildings. In contrast, the empirical model treats a single building as a unit and assesses the energy-saving benefits of green roofs based on the conversion of air temperature reduction and heat absorption in greenspaces\cite{ZHANG201437}. While the empirical method may yield lower accuracy compared to simulation methods, it facilitates a preliminary assessment at an urban scale.

In the process of urbanization, increased anthropogenic activity, such as the burning of coal, natural gas, and oil, is contributing to the emission of carbon dioxide, resulting in climate change and global warming. To address the climate crisis, countries committed to achieving net-zero emissions by 2050 or earlier as part of the 2015 Paris Agreement. One potential solution to contribute to these goals is the implementation of green roofs. However, the results reveal that green roofs play a small role in Hong Kong's annual carbon emission offsets. An intriguing finding is that carbon emission reduction indirectly achieved by the energy-saving benefits of green roofs is about twice that of the direct pattern, while green roofs only reduce the air temperature by about 0.15\degree{C} (Fig.~\ref{fig:4}). Consequently, a higher degree of energy savings could lead to a greater of annual carbon emission offsets. Rooftop photovoltaics have emerged as an effective solution, offering both thermal insulation and the conversion of solar radiation into electrical energy, thereby holding significant promise for decarbonizing the power sector\cite{zhang2023carbon}. In this context, the integration of rooftop photovoltaics and green roofs presents an opportunity to play a more substantial role in achieving climate goals\cite{shafique2020photovoltaic, FLECK2022109703}. Therefore, assessing the comprehensive benefits of the integration of these two roof technologies at an urban scale is a topic worthy of exploration in future research on sustainable urban development.

\begin{figure}[h]
\centering
\includegraphics[width=1\linewidth]{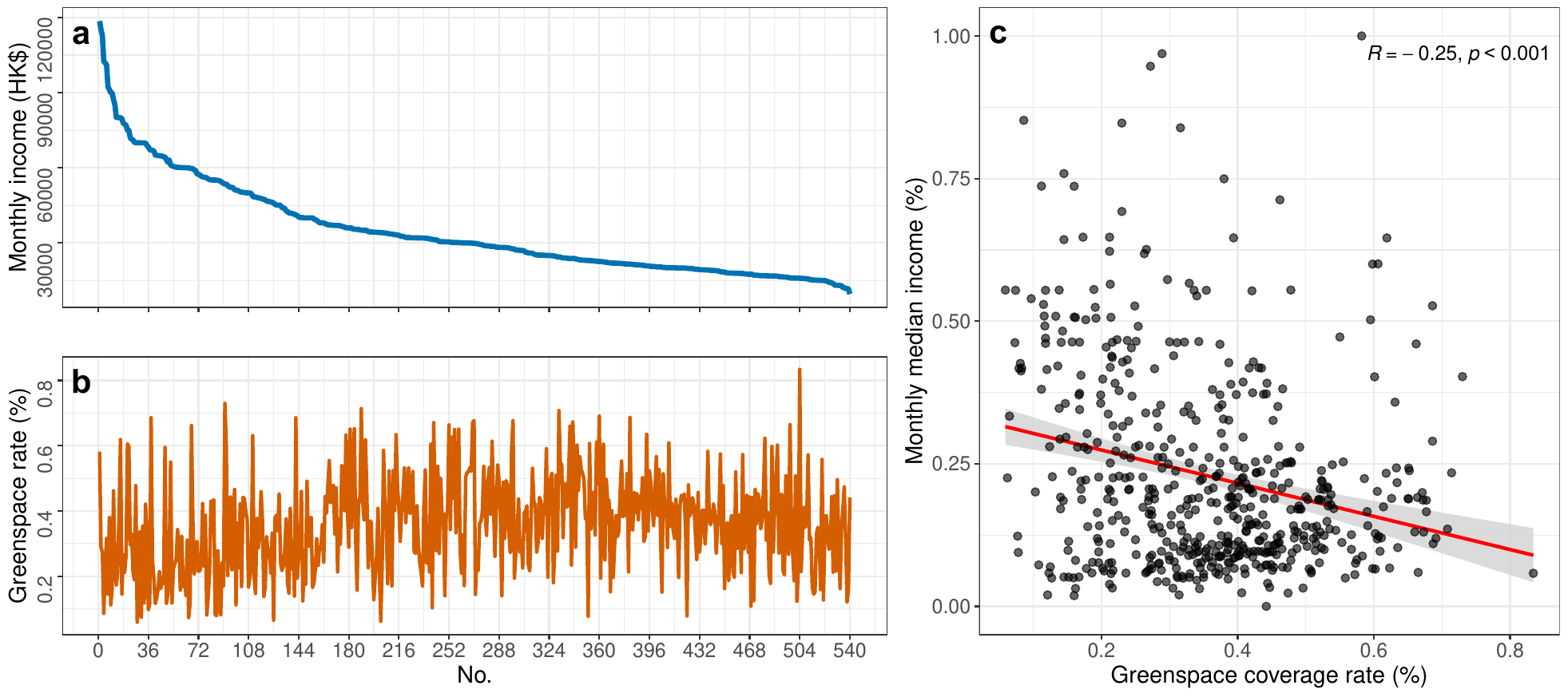}
\caption{\textbf{$\vert$ The correlation (c) between monthly income (a) and greenspace coverage rate (b) of buildings.} \textbf{a}, The income data used in this analysis is derived from the monthly median income of 540 housing estates in 2021, according to Hong Kong government statistics. \textbf{b}, The greenspace coverage rate refers to the proportion of greenspace coverage area within a 500-meter horizontal radius of the building.}
\label{fig:5}
\end{figure}

The relationship between humans and greenspaces has garnered increased attention in the field of social economics\cite{GREENE201824}. Many studies have found inequality in greenspace exposure for different groups. Some researchers argue that advantaged residents, such as those with higher incomes and education levels, tend to have greater exposure to abundant greenspaces\cite{FERGUSON2018136,yin2023unequal}. However, some studies have yielded contradictory conclusions\cite{LIU2017130, XIAO2017383}. These findings have implications for the assessment of green roofs. To assess the socioeconomic benefits in Hong Kong, we analyzed the correlation between income and greenspace coverage rate. Fig.~\ref{fig:5}a shows a significant disparity in the median monthly income among different groups, with the richest earning ten times more than the poorest. In contrast, the overall greenspace coverage rate among these groups was relatively equal (Fig.~\ref{fig:5}b). Consequently, we observed a significant weak negative correlation between income and greenspace coverage rate (R=-0.25, p\textless0.001; Fig.~\ref{fig:5}c). This suggests different groups enjoy relatively equal exposure to greenspace services in Hong Kong, but indirectly indicates that green roofs have a limited effect on improving the income level of Hong Kong residents.

As urbanization advances, the harmonious coexistence of people and nature within cities faces significant pressure\cite{cuthbert2022global}. Roof greening is considered an effective strategy to alleviate this issue\cite{SHAFIQUE2018757}. Consequently, rational roof greening has become a critical challenge for city planners. Data serves as the foundational element in tackling this problem. Previous studies typically relied on less precise and coarser-resolution data\cite{XU2021108392, LIU2023128920} and focused on regional or zone-level rough analysis\cite{HONG2019102468}. The geospatial big data utilized in this study allows for a city-wide evaluation of green roofs at an individual building level, which offers solutions for reconciling the contradiction between large-scale and refined analysis in urban processes. Furthermore, cities are complex ecosystems comprising nature, society, and economy. Therefore, assessing roof greening priority based on a single factor might introduce bias\cite{XU2021108392, VIECCO2021108120}. The proposed framework for assessing roof greening priority, which considers environmental, socioeconomic, and climate factors, appears more balanced and accurately reflects the rich ecological connotation of cities. Similarly, relying on a single-benefit assessment may lead to a subjective understanding of the role of green roofs in urban processes\cite{SHAFIQUE2020119471}. For instance, roof greening can greatly increase the greenspace coverage rate around buildings and the greenspace exposure in Hong Kong, while playing a small role in urban heat mitigation and carbon neutrality. Accordingly, a multifaceted assessment is necessary to objectively understand the role of green roofs in urban processes. Currently, basic geographical datasets in many cities are continually being improved and accessible, with various technologies, including deep learning, gradually gaining popularity. These advancements enable the proposed framework for assessing green roofs and its findings to hold the potential for transferability to other cities of a similar nature. In summary, this study provides unprecedentedly high precision and spatial-fidelity insights on urban planning and policy-making, from the data to the methods and the outcomes. These insights are valuable for contributing to the advancement of sustainable development in cities worldwide.

\section{Methods}\label{sec4}

\subsection{Research design}\label{sec41}

To promote urban resilience and sustainability in Hong Kong, a detailed flowchart for assessing green roofs was presented in Supplementary Fig. 5. This flowchart involved the integration of airborne laser scanning point clouds, Google Maps data, and building census data to gather building attributes such as roof slope and area, building age and category, which were used to identify potential green roofs. Subsequently, six greening indicators were designed from a sustainable urban development perspective, and satellite remote sensing images, road network data, weather data, and government statistical data were supplemented to quantify them. Then, all indicators were incorporated to evaluate the greening priority. Furthermore, the benefits of green roofs on the environment, urban heat mitigation, carbon emission reduction, and social economy, were assessed through simulation and estimation.

\subsection{Environmental data}\label{sec42}

The urban geospatial big data, comprising airborne laser scanning point clouds, geographic data of building footprints and road networks, Google Maps data, satellite images, annual precipitation, government statistics, and population data were utilized to validate the feasibility of this study. Airborne laser scanning data with centimeter-level resolution were employed to extract roofs and calculate roof slope and area, building height, and greenspace coverage rate. Geographic data of building footprints and road networks were utilized to extract building attributes and analyze the distance between each building and its nearest main road. Google Maps data with a resolution of 0.8 m were used to predict building age and category. For satellite data, Landsat 8 provided a total of four multispectral images with a resolution of 30 m, covering the period from March 2021 to February 2022. These images were used to retrieve the surface temperature of the four seasons. Annual precipitation data for 2021, collected from 139 weather stations, were used to generate the precipitation distribution across the entire territory. Median monthly income data for 2021, released by the government, were utilized to quantify the economic indicator. This income data was derived from 540 housing estates and all districts. Lastly, population data from the WorldPop dataset with a resolution of 100 m, corresponding to the year 2020, were used to calculate human exposure to greenspace index value.

\subsection{Extraction of building attributes}\label{sec43}

The physical attributes of the roof are crucial metrics for assessing the feasibility of roof greening. Combining the implementation specifications and effects of roof greening, geometric attributes including roof slope and area, as well as building age, were used to identify potential green roofs in this study. Notably, all roofs detected from airborne laser scanning point clouds were used as the basis to couple multi-source data. Building points were projected onto the horizontal plane, where the highest point at each position was identified as a candidate roof point. To filter out the edge and wall points, we compared the elevation difference between the candidate roof point and its 4 neighboring points. If the difference was less than 1 m, the point was considered a roof point. After detecting all roof points, the connected component labeling algorithm was used to cluster them, followed by segmenting roofs using the region growing algorithm\cite{SHAO2021103660}. The slope and area of each segment were then calculated, in which segments with a slope less than 15 degrees and an area greater than 10 m$^2$ were considered as potential green roofs (Supplementary Fig. 6). Furthermore, to prioritize green roofs, building age and category were identified as critical factors. However, the information on these factors was lacking for many buildings. A deep learning technique was employed to predict the age and category of these buildings by creating super-resolution images. The geographic data of building footprints was first fused with Google Maps data based on geo-referenced coordinates. Subsequently, the top-view image for each building was cropped from Google Maps data based on the building footprint. Finally, the images of buildings with known age and category information were used as training samples to predict the age and category of other buildings using the deep learning network SR3\cite{Saharia9887996} (Supplementary Fig. 7 and Table 1). Buildings older than 60 years were considered to have inadequate load-bearing capacity and were deemed unsuitable for roof greening.

\subsection{Measurement of indicator and priority}\label{sec44}

\textbf{Measurement of greening indicator.} Based on the environmental, socioeconomic, and climate factors concerning the sustainable urban development goals, six indicators, including the greenspace coverage rate, the distance of building to road, building category, resident income, surface temperature, and annual precipitation, were designed to evaluate roof greening priorities in this study.

The relationship between buildings and the surrounding greenspaces plays a crucial role in determining the urgency of roof greening. Generally, buildings that are located far away from green spaces, such as parks, require green roofs more than those surrounded by green spaces\cite{XU2021108392}. Additionally, due to the benefits of green roofs in absorbing vehicle exhaust and noise, buildings located closer to main roads typically require more greenery\cite{han2022surrounding}. Thus, the greenspace coverage rate around buildings and the distance of the building to the road were selected as two pertinent environmental indicators. Specifically, we projected all potential roof points and vegetation points onto the horizontal plane, resulting in a map with a resolution of 5 m. Using this map, the greenspace coverage rate within a 500-meter radius of each roof point was calculated by Eq. (\ref{equ:gc})\cite{chen2022contrasting}. The average rate of all points on a roof was then determined as the final greenspace coverage rate for this roof.
\begin{equation} \label{equ:gc}
GC=\frac{\sum_{i=1}^{n}G_i}{A}
\end{equation}
where $G_i$ represents the size (5 × 5 m$^2$) of a pixel, $n$ is the total number of pixels that were covered by vegetation, $A$ represents the area of a circle with a radius of 500 m, and $GC$ is the greenspace coverage rate of a roof point. The horizontal distance of the buildings to the nearest main road was used to quantify the distance indicator. As the distance increases, the impacts of vehicles diminish, resulting in a decrease in the demand for greening and the distance indicator value. When the distance exceeds 500 m, the distance indicator value is set to 0.

In terms of social economics, building category and resident income were used to assess roof greening priority. The category indicates the building ownership, including private (such as commercial buildings), public (such as government offices, public schools, and hospitals), and miscellaneous buildings (including temporary and open structures). Private building owners may be less motivated to invest in green roofs due to the high costs involved, so the greening indicator value for private buildings was set at 0.5. Conversely, public building owners are more likely to actively embrace sustainable development policies, so the greening indicator value for these buildings was set at 1. The miscellaneous category falls between the private and public categories, with an indicator value of 0.75. Moreover, previous studies have suggested a correlation between greenspaces and economies\cite{chen2022contrasting}, such as per capita income. To quantify the economic indicator, we combined the median monthly income data from 540 housing estates and all districts and then employed the inverse distance weighted algorithm to obtain the income distribution for the year 2021.

Given the hot and rainy climates in Hong Kong, surface temperature and annual precipitation were included as two greening indicators. Landsat 8 thermal infrared and multispectral data were combined in the Google Earth Engine platform to derive surface temperature at a resolution of 30 meters\cite{rs12091471}. Typically, a clear-sky image was selected for the inversion of surface temperature. However, satellite images often have limitations during the summer and autumn seasons due to cloud cover. Thus, we manually removed cloud-occluded areas from images and used the Kriging interpolation algorithm to fill in missing temperature data within those regions. Due to Hong Kong’s climate characteristics, we assigned March to May as spring, June to August as summer, September to November as autumn, and December to February as winter. Furthermore, considering the low resolution of existing precipitation remote sensing data, Kriging interpolation on ground records was performed to obtain the distribution of annual precipitation at a resolution of 30 meters.

\vspace{1em}

\noindent\textbf{Measurement of greening priority.} The process of using the aforementioned six indicators to measure roof greening priorities is a typical weight assignment problem. Common methods include the entropy method, variance coefficient method, CRITIC (Criteria Importance Through Inter-criteria Correlation) method, and equal weighting method\cite{XU2021108392, ZHANG2021106762}. The first three methods assign weights to each indicator based on the amount of information or correlation between indicators, belonging to the objective weighting method. The equal weighting method directly assigns the same weight to each indicator. According to the comparison, the equal weighting method appears to be more reasonable and in line with the principle that each greening indicator has equal importance in urban sustainability (refer to Supplementary Table 2). Moreover, due to the variations in temperature across different seasons, we combined the temperature data from all four seasons to derive the temperature indicator value. Given the varying cooling demands, a subjective weighting method was utilized to assign weights to each season. Recognizing the urgency for cooling during summer and autumn, these seasons were assigned a higher weight (0.4). Conversely, spring and winter, which require less cooling, were assigned a lower weight of 0.1. The comparison reveals that this method is more reasonable than the objective weight method (refer to Supplementary Table 3). Consequently, roof greening priority can be calculated by:
\begin{equation} \label{equ:P&I}
\begin{aligned}
    & P=\frac{1}{6}\left(I_g+I_d+I_c+I_i+I_t+I_p\right);\ \left\{0\le P,I_g,I_d,I_c,I_i,I_t,I_p\le1\right\} \\
    & I_t=\frac{1}{10}\left(T_1+4T_2+4T_3+T_4\right);\ \left\{0\le T_1,T_2,T_3,T_4\le1\right\}
\end{aligned}
\end{equation}
where $P$ is the final priority of roof greening, $\left\{I_g,I_d,I_i,I_t,I_p\right\}$ are nondimensionalized prioritizing indicators, representing the greenspace coverage rate $I_g$, the distance between buildings and the nearest main road $I_d$, the median monthly income $I_i$, the surface temperature $I_t$, and the annual precipitation $I_p$. $I_c$ represents the building category indicator. $T_1$, $T_2$, $T_3$, and $T_4$ represent the nondimensionalized temperatures of spring, summer, autumn, and winter, respectively. During the nondimensionalization process of all indicators, $I_t$ and $I_p$ were considered as positive indicators, while $I_g$, $I_d$, and $I_i$ were treated as negative indicators.

\subsection{Benefit evaluation of green roofs}\label{sec45}

The conventional indicator of greenspace coverage often fails to consider the accessibility to the population\cite{SONG2021106778}, making it challenging to quantify the connection between humans and nature. To address these limitations, the concept of human exposure to greenspace index was introduced, which aimed to capture the level of residents’ exposure to nature in the form of greenspaces. This term does not solely refer to human visits to green spaces but also encompasses environmental, socio-economic, physical, and mental health benefits\cite{chen2022contrasting, wong2021greenery}. The index value was calculated by combining a population-weighted model with Eq. (\ref{equ:gc})\cite{chen2022contrasting}.

To evaluate the effectiveness of green roofs in mitigating urban heat, WRF/UCM, a sophisticated numerical weather prediction system\cite{skamarock2019description}, was used to simulate the air temperature at 2 m above the ground under bare roof and green roof conditions. WRF is a mesoscale model applied in meteorology and air quality analysis. To provide a more accurate description of the urban surface process, the WRF model was coupled with the single-layer UCM. UCM is embedded within the first model layer and considers detailed canopy processes, including the shadowing, reflections, and trapping of radiation, as well as the surface energy budget of roofs, walls, and roads, and anthropogenic heat emission\cite{KUSAKA2001}. WRF/UCM has been widely used in simulating urban heat islands and developing strategies to mitigate them\cite{WANG2022109082}. Specifically, the model was operated with two nested domains at resolutions of 300 m and 100 m, respectively. The first domain covers the Guangdong-Hong Kong-Macao Greater Bay Area, and the second domain covers Hong Kong. The urban land use was determined using the default dataset of the WRF model. We used 38 vertical levels from the surface to 50hPa (top of the atmosphere), with levels more closely spaced near the surface, to better resolve near-surface atmospheric processes. The physical parameterization schemes used in this study are as follows: WRF Single-Moment 3-class simple ice scheme for Microphysics; RRTM scheme for long-wave radiation; Dudhia shortwave scheme for short-wave radiation; MM5 Monin-Obukhov scheme for surface clay physics; Yonsei University scheme for planetary boundary layer physics; the Noah-LSM for natural land surface processes; and UCM for urban surface processes\cite{IMRAN2018393, ARGHAVANI2020121183}. Additionally, given the time complexity, the air temperature simulation process in this study only focused on four representative sunny days from January 15 (winter), April 13 (spring), July 14 (summer), and October 2 (autumn) in 2021.

The combination of field surveys and remote sensing inversion is a common method for estimating large-scale carbon sequestration. However, roof greening has not yet been implemented, so remote sensing monitoring technology may not be applicable. Using empirical values for inference is an alternative method. This type of method is simple and does not involve complicated processes and models, with the determination of carbon sequestration per unit area of green roofs being crucial. Given the complexity of field surveys, the measurement based on literature-records values is often used to determine the above parameters\cite{DONG2023113376}. Plant photosynthesis and energy conservation are two important patterns through which green roofs reduce carbon emissions\cite{SHAFIQUE2020119471, ZHENG2023128018}, in which the carbon emissions directly sequestered through plant photosynthesis can be empirically estimated in Eq. (\ref{equ:C_v}):
\begin{equation} \label{equ:C_v}
    C_v=Q_{co2}\times A_r
\end{equation}
where $C_v$ is carbon dioxide weight absorbed by green roofs in one year (kg CO$_2$$\cdot\text{yr}^{-1}$), $Q_{co2}$ is annual carbon sequestration per unit area of green roofs (kg CO$_2$$\cdot(\text{m}^2\cdot\text{yr})^{-1}$). Indigenous herbs were considered to be the most suitable plant species for roof greening in Hong Kong, with extensive green roofs being the predominant type. According to the literature-recorded values\cite{KARTERIS2016510, KURONUMA2017, su10072256, CAI201919, SEYEDABADI2021100119} (refer to Supplementary Table 4), we empirically assumed $Q_{co2}$ to be 1.46 kg CO$_2$$\cdot(\text{m}^2\cdot\text{yr})^{-1}$.

Green roofs can save energy through urban heat mitigation, thereby indirectly benefiting carbon emission reduction. In this study, an empirical model\cite{ZHANG201437} was utilized to assess the energy-saving benefits on an urban scale by examining the conversion of cooling amplitude and heat absorption in green spaces. This model considered a theoretical air column as a computational unit. When the plants in this air column absorb the heat, the surrounding air temperature decreases by $\Delta T$. Consequently, the mitigated heat $\Delta E$ (J$\cdot$(m$^{3}$$\cdot$h)$^{-1}$) can be determined by the temperature reduction and the volumetric heat capacity of the air ($\rho_{c}$) in Eq. (\ref{equ:E}):
\begin{equation}
\label{equ:E}
   \Delta E=\Delta T \times \rho_{c}
\end{equation}
where $\rho_{c}$ (J$\cdot$(m$^{3}$$\cdot$\degree{C})$^{-1}$) is calculated by multiplying the specific heat capacity $c_{air}$ (1004 J$\cdot$(kg$\cdot\degree$C)$^{-1}$) by the air density $d_{air}$ (1.29 kg$\cdot$m$^{-3}$). $\Delta T$ is the average temperature reduction per unit hour (\degree{C}$\cdot$h$^{-1}$). Generally, the cooling effect of green roofs varies depending on the weather conditions, in which the effect is the most significant on sunny days, followed by cloudy days, but it is almost negligible on rainy days. Previous studies have shown that on cloudy days, the cooling load reduction of extensive green roofs is about 0.63 times that of sunny days in Hong Kong\cite{JIM20129}. Therefore, according to the simulation results (Fig.~\ref{fig:4}), $\Delta T$ is set to 0.15\degree{C} for sunny days and 0.1\degree{C} for cloudy days. Notably, Eq. (\ref{equ:E}) only calculates the heat consumed by cooling per unit air column in one hour. Building volume and cooling time should be considered for evaluating annual energy-saving benefits. In this study, each building was assumed to be a rectangular air column, and then the annual energy $E$ (J) saved by green roofs in Hong Kong can be calculated by:
\begin{equation} \label{equ:C_e}
    E=\Delta T\times c_{air}\times d_{air}\times t\times \sum_{i}^{N}{(A_{i}\times h_{i})}
\end{equation}
where $A_{i}$ (m$^2$) and $h_{i}$ (m) (Supplementary Fig. 4) are the area and height of a building extracted from airborne laser scanning point clouds, and $N$ is the number of buildings. $t$ is the time (h) that electrical power needs to be used for cooling during summer and autumn (180 days). Statistically, there are 30 rainy days, and assuming the partitioning of sunny and cloudy days is 1:1 throughout the summer and autumn of 2021. Furthermore, the carbon emissions indirectly reduced by energy saving can be estimated based on the conversion between them:
\begin{equation} \label{equ:C_e}
    C_e=E\times k
\end{equation}
where $C_e$ (kg CO$_2$$\cdot\text{yr}^{-1}$) is the annual equivalent amount of carbon emission converted by energy saving, and $k$ is the conversion coefficient, representing the carbon emissions produced by 1 kW$\cdot$h of electricity, which is set to 0.785 kg CO$_2$$\cdot$(kW$\cdot$h)$^{-1}$\cite{ZHANG201437}. Thus, the total carbon emission reduction is calculated by adding $C_v$ and $C_e$.

According to the carbon trading and electricity prices in Hong Kong at the time of the study, the financial gains related to carbon emission reduction and energy savings were estimated to assess the economic benefits of roof greening. Furthermore, a linear regression analysis was performed to examine the correlation between greenspace coverage rate and median monthly income. By discussing this relationship, we explored the fairness of access to greenspaces and the potential of green roofs to improve the living standards of disadvantaged residents.


\section*{Data availability}

The geographic big data used in this study can be downloaded from the following sources: Airborne laser scanning point clouds and the geographic information data related to building footprint and road data can be obtained from the Common Spatial Data Infrastructure Portal (\url{https://portal.csdi.gov.hk/csdi-webpage/}); Google Maps data was obtained by using QGIS software from the tile URL (\url{http://mt0.google.com/vt/lyrs=s&hl=en&x=%7bx%7d&y=%7by%7d&z=%7bz%7d}); Satellite remote sensing images are from Landsat 8 (\url{https://landsat.gsfc.nasa.gov/satellites/landsat-8/}); A priori building age information can be obtained from the Hong Kong private building dataset (\url{https://initiumdata.carto.com/me}); The data on annual precipitations is collected from various weather stations (\url{https://i-lens.hk/hkweather/}); The population data is obtained from the WorldPop (\url{https://www.worldpop.org/}); The data on median monthly income is sourced from the 2021 Population Census (\url{https://www.census2021.gov.hk/}). Some input data have been deposited at the repository: \url{https://polyuit-my.sharepoint.com/:f:/g/personal/zhiyihe_polyu_edu_hk/EkCaXXlvoBtHgfYAy_t2PMABXHusVrbMnAlIZ-64Peo_jg?e=OoqGIH}.

\section*{Code availability}
The codes used for processing data (including point clouds and images) and analyses are available at \url{https://github.com/schojer/gr}.

\bibliography{sn-bibliography}

\backmatter

\section*{Acknowledgments}

This work was supported by the National Natural Science Foundation of China (Grant No.42171361 and Grant No.42301432) and the Research Grants Council of the Hong Kong Special Administrative Region, China, under Project PolyU 15215023. This work was also funded by the research project (Project Number: 2021.A6.184.21D) of the Public Policy Research Funding Scheme of The Government of the Hong Kong Special Administrative Region. This work was partially supported by The Hong Kong Polytechnic University (Projects 1-YXAQ and Q-CDAU) and the Innovative Research Program of International Research Center of Big Data for Sustainable Development Goals (Grant CBAS2022IRP09).

\section*{Author contributions}

J.S, W.Y and H.G conceived the idea. L.L, L.Z, Z.H and P.W prepared the data. J.S, W.Y, L.L, L.Z and Z.H developed the theoretical models and processed data. J.S, L.L, L.Z, Z.H and P.W performed the experiments and analyzed the data. All authors contributed to the discussion and editing of manuscripts.

\section*{Competing interests}
The authors declare no competing interests.

\section*{Additional information}
\textbf{Supplementary information} contains Supplementary Figs. 1 to 7 and Supplementary Tables 1 to 4.

\end{document}